\documentclass[conference]{IEEEtran}
\IEEEoverridecommandlockouts
\usepackage{cite}
\usepackage{amsmath,amssymb,amsfonts}
\usepackage{algorithmic}
\usepackage{graphicx}
\usepackage{textcomp}
\usepackage{xcolor}
\usepackage{booktabs}
\newcommand{\nj}[1]{\textcolor{black}{#1}}
\newcommand{\ms}[1]{\textcolor{black}{#1}}
\newcommand{\jy}[1]{\textcolor{black}{#1}}

\def\BibTeX{{\rm B\kern-.05em{\sc i\kern-.025em b}\kern-.08em
    T\kern-.1667em\lower.7ex\hbox{E}\kern-.125emX}}
\begin{document}

\title{Task-oriented Design through Deep Reinforcement Learning\\
\thanks{\IEEEauthorrefmark{1} equal contribution}
}

\author{\IEEEauthorblockN{Junyoung Choi\IEEEauthorrefmark{1}, Minsung Hyun\IEEEauthorrefmark{1}, Nojun Kwak}
\IEEEauthorblockA{\textit{Department of Transdisciplinary Studies} \\
\textit{Seoul National University}\\
Seoul, Korea Republic of. \\
\{djcola814, minsung.hyun, nojunk\}@snu.ac.kr}
% \and
% \and
% \and
% \and
% \and
}

\maketitle

\begin{abstract}
We propose a new low-cost machine-learning-based methodology which assists designers in reducing the gap between the problem and the solution in the design process. Our work applies reinforcement learning (RL) to find the optimal task-oriented design solution through the construction of the design action for each task. For this task-oriented design, the 3D design process in product design is assigned to an action space in Deep RL, and the desired 3D model is obtained by training each design action according to the task. 
By showing that this method achieves satisfactory design even when applied to a task pursuing multiple goals, we suggest the direction of how machine learning can contribute to the design process. Also, we have validated with product designers that this methodology can assist the creative part in the process of design.
%본 논문은, design process에서 problem 과 solution사이의 거리를 줄이는데 designer들에게 도움을 줄 수 있는 새로운 low-cost computer-based methodology를 제시한다. 본 연구에서는, design research로부터 도출된 classic problems or task를 해결하기 위해, 각각의 task에 대한 design solution의 방향성을 deep reinforcement learning algorithm을 통해 제시한다. 이들은 각각의 task에 대한 design action의 설계를 통해 얻어진 optimal task-oriented design이 될 것이다. 이러한 task-oriented design을 위해 product design에서의 3D design process 를 Deep-RL내의 action space에 할당하고, 각각의 design action을 task에 맞춰 학습시킴으로써 원하는 디자인을 얻어냈다. 또한 이 방법을 다양한 목표를 추구하는 task에 적용할 경우에도 만족스러운 디자인을 얻어낼 수 있다는 것을 보임으로써 design process에서 machine learning이 기여할 수 있는 새로운 methodology의 방향성을 제시한다. (또한, 이를 다양한 task에 대해 수행하여, 두가지 이상 혹은 심지어 두 task가 일부 상충되는 경우에서도, 각각의 task를 모두 수행할 수 있는 새로운 디자인을 굉장히 쉽게 얻어낼 수 있다는 것을 보임으로써, design process내에서 machine learning이 기여할 수 있는 새로운 methodology의 방향성을 제시한다.)
\end{abstract}


\section{Introduction}
\begin{figure*}
	\centering
	\includegraphics[width = 0.8\linewidth]{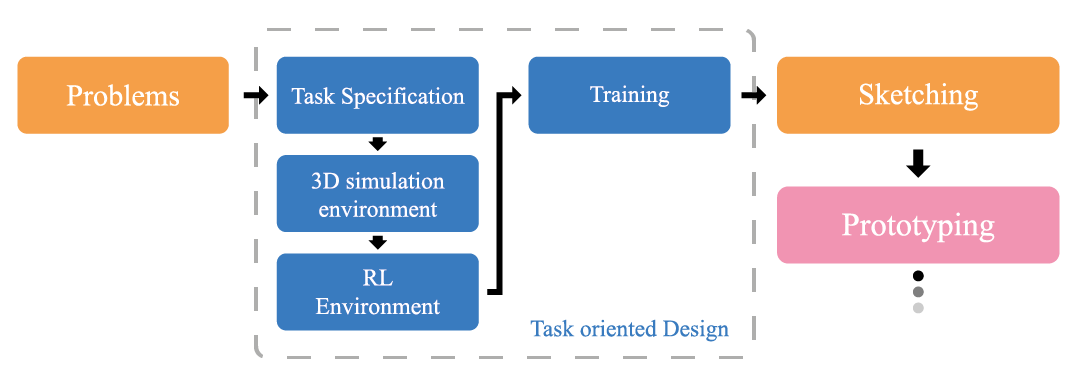}\\
	\caption{The proposed method of task-oriented design} through deep reinforcement learning in \nj{the} product design process. \jy{With task specification, 3D simulation and a reinforcement learning environment, our methodology propose a methodology that \ms{directly assists creative part of }%minimizes user involvements in 
    the product design process.}
	\label{figure_tod}
\end{figure*}
This paper suggests a methodology of helping designers to obtain morphological insights with the use of  deep reinforcement learning. Within the constructive design process, design methodologies have evolved to find the points of human inconvenience or the problems contained in the use of the product ~\cite{DRTP2011}. %After going through problem statement process, %another 
In the problem solving process, a design solution for the defined problem is searched for that can be applied to the actual design. Many designers have suggested various methods to solve these design problems \cite{Dorst2001, MDP1991}. However, since these methods usually require high cost in terms of time and space, many researchers have difficulties in finding a good solution. 

%제품 디자인 프로세스 내에서, UI/UX 디자인은, 인간의 불편함과 제품을 사용함에 담겨있는 문제를 찾기 위한 방향으로 발전되어왔다. 이를 위해 여러 User research technique을 이용하여, 실생활에 산재되어 있는 문제들을 찾아내는 것이 현재 디자인 디자인 산업에서 핵심적인 방향성으로 제시되어 왔다. 그리고 위의 과정을 통해 problem statement 과정을 거치면, 찾아진 problem을 활용하여 이를 해결할 수 있는 solution을 실제 디자인에 적용할 수 있는 Design solution을 찾는 것이 또 다른 목표이다. 이 과정에서, 많은 디자이너들은 이 problem과 solution을 이을 수 있는 많은 방법론들을 제시해왔다.\cite{Dorst2001, MDP1991} 하지만 이들의 경우 보통 시간적/공간적으로 high-cost를 요구하였기 때문에, 좋은 problem을 찾았음에도 불구하고, 이를 통해 좋은 solution을 찾는것에서 많은 연구자들이 고통을 겪었다.

Meanwhile, the role of computer science in the design process has been changed. Computer scientists collaborate not only in the development of design tools that designers can use directly for designing products, but also in various forms that help designers with computer simulation or machine learning algorithms before the mock-up stage \cite{Aish2009, du2016, parish2001, umentani2012, Zhu2015}. However, these approaches only use computer science as a rule-based assistant, not a creative designing tool. Dreamsketch~\cite{kazi2017}, on the other hand, suggested the methodology to obtain multiple 3D design solutions based on the context created from the sketch phase. But still, because it is applied only after all the analysis has been done, rather than used for understanding the context of design, it is of little help in the design process and does not directly address the underlying problem solving. Pahn et al. \cite{Pahn2001} is another good example that attempted to assist the design process using machine learning, but it is very costly because it requires the process of dividing the product and the corresponding design problem into small objects and assigns a role to each object.

%반면, Pahn et al.\cite{Pahn2001}의 경우 design process에 직접적으로 도움을 주는 역할이긴 하나, product를 각각 object로 쪼개 하나하나의 역할을 부여해주는 방식이기 때문에 그 과정이 굉장히 high-cost이다.
%한편, 최근 computer science가 디자인 프로세스 내에서 하는 역할이 변하고 있다. Computer scientist 들은 product를 직접적으로 디자인하는 3D tool의 발전 뿐만 아니라, 디자인 단계에서 computer simulation을 통해 디자인에 도움을 받는 형식으로 협업하고 있다. \cite{Aish2009}\cite{du2016}\cite{parish2001}\cite{umentani2012} Dreamsketch\cite{kazi2017}에서는 product design process 의 sketch 단계로부터 만들어진 context를 기반으로 multiple 3d design solution을 얻어내었다. But still, design process의 creative part에서의 이용보다는, 모든 analysis를 끝낸 이후 design process에서 약간의 도움을 받는 형식으로, 근본적인 문제 해결에 직접적으로 관여하지는 않는다.
Different from the above mentioned methods, we have chosen algorithms that can more directly understand the design process in order to suggest a more direct and lower-cost methodology to designers at the problem-solution bridge stage. The reinforcement learning algorithm, which is becoming a hot topic in computer science, is an algorithm that learns the best action (to get the best reward) in a provided environment. Even if we do not give detailed information about the intermediate process leading to the best reward, we can learn to get the best action possible. Through this, we devised a framework that can design based on a given task.
%우리는 위에 이야기 했던 근본적인 problem - solution사이의 거리를 좁히기 위해 기존 방식보다 좀 더 직접적으로 design process에 영향을 미칠 수 있는 알고리즘을 선별했다. Computer science에서 화두가 되고 있는 Reinforcement Learning algorithm은, 어떠한 environment가 제공되었을 때 이 environment에서 최선의 행동을 하도록 (최고의 reward를 얻도록) 학습하는 알고리즘이다. 최고의 reward를 향해 가는 중간 과정에 대한 정보를 세세히 알려주지 않는다고 하더라도, 최선의 action을 얻을 수 있도록 학습되는 것이다. 이를 통해 aesthetic 뿐만 아니라 task에 기반하여 design을 할 수 있는 알고리즘 (or framework??)을 고안했다.

In this paper, we apply the reinforcement learning algorithm to product design and present a link that enables the computer to directly find the task-oriented solution through the problem. The whole proposed methodology is shown in Figure \ref{figure_tod}. When problems from design research is given, we define tasks, \ms{3D simulation} environments and \ms{reinforcement learning environments.} %reward function which are necessary for \nj{a} reinforcement learning.
Tasks and 3D simulation environments are processed in \textsc{Blender} \cite{Blender} and linked to the reinforcement learning algorithm. % with defining reward function. 
\nj{In the figure, this paper deals with the processes denoted in the \ms{dashed-line} box. } 

Zhu et al. \cite{Zhu2015} is also a good example of a deep learning algorithm for product interpretation. However, we go \nj{one} step further and present a methodology that allows the computer to design itself based on the understanding of the product or task. \nj{More specifically, we have chosen to design a pot with a couple of design objectives. By tackling this problem,} we will discuss how the reinforcement learning algorithm finds solutions in order to achieve high scores in \nj{a given task}. \nj{To enable this}, we define the design process as an action space that can be understood by the computer. 
%\jy{\sout{in order to apply the traditional deep reinforcement learning algorithm to the design}}. 
We will also cover how we can finally use the \jy{\nj{generated} output by giving morphological intuition to product designers.
}

\section{Related Work}

\subsection{Constructive Design}
In various study of product design process \cite{DRTP2011, Bang2012}, the \nj{authors} discuss about \nj{the process of} constructive design research \nj{which} is initiated by formulating a research question out of an existing theory or philosophy, \nj{then investigate the question} through a process of making and designing artifacts. 

For constructive design, user research should be proceeded first. With studying users and products, designers get several insights that should be applied to their final design. After studio work, constructive design researchers \nj{develop designs,} %development, 
which begins with sketchy ideas and mock-ups. In this stage, usually hundreds of mock-ups are made by designers, which \nj{costs} a lot of time and efforts.

\subsection{Reinforcement Learning}
Reinforcement learning is an algorithm that learns which actions to take to maximize rewards in a given environment. In reinforcement learning, we define and solve problems with \nj{the framework of Markov Decision Processes (MDPs)}. MDPs consist of the environment and the agent. They interact each other at every \nj{continual} time \nj{index} $t = 0,1,2, \cdots$, 
%as time passes 
and the agent tries to achieve a given goal.
 %강화학습은 주어진 환경에서 보상을 극대화하기 위해 어떤 행동을 취해야 할지를 배워나가는 알고리즘이다. 강화학습에서는 문제를 Markov Decision Processes(MDPs) Framework로 정의하고 해결한다. MDPs는 환경과 주어진 목표를 달성하려는 에이전트의 상호작용으로 구성된다. 환경과 에이전트는 시간의 흐름에 따라 매 $t=0,1,2,3, \cdots$ 시점마다 상호작용하는데, 구체적으로 설명하면 다음과 같다.  

Specifically, the agent receives information about the state from the environment and takes action to obtain maximum rewards in the current state, and the action is determined by the policy. The policy is defined as a probability distribution ($\pi(a|s)$) that is of available action outputs in a given state \cite{sutton2017reinforcement}. The environment outputs rewards as a result of agent's action and this process repeats. In the end, trajectories such as $s_0,a_0,r_1,s_1,a_1,r_2,s_2,a_2,r_3,\cdots$ can be obtained by the interaction of \nj{the agent and the environment}. State changes are determined by stationary transition dynamics distribution $p(s_{t+1}|s_t,a_t)=p(s_{t+1}|s_0,a_0,\cdots,s_t,a_t)$ which follows markov properties \cite{silver2014deterministic}.
%에이전트는 환경으로부터 상태에 대한 정보를 받아 현재 상태에서 가장 큰 보상을 얻을 수 있는 행동을 취하며, 행동은 정책에 의해 결정된다. 정책은 주어진 상태에서 가능한 행동을 출력하는 확률분포 $\pi_{\theta}(a|s)$로 정의된다.\cite{sutton2017reinforcement} 에이전트의 행동에 대한 결과로 다시 환경에서는 보상값을 출력하고 이 과정이 반복된다. 결과적으로 환경과 에이전트의 상호작용으로 인해 $s_0,a_0,r_1,s_1,a_1,r_2,s_2,a_2,r_3,\cdots$와 같은 경로(trajectory)를 얻을 수 있으며 행동에 따른 상태의 변화는 마르코프 특성을 따르는 stationary transition dynamics distribution $p(s_{t+1}|s_t,a_t)=p(s_{t+1}|s_0,a_0,\cdots,s_t,a_t)$에 의해 결정된다.\cite{silver2014deterministic} 

Deep Q-Network(DQN) has achieved surprising performance in the Atari2600 task learning environment, where the action-value function ($q(s_{t},a_{t})$; Q-function) \ms{was approximated by deep neural network \cite{mnih2013playing}. The Q-function estimates the maximum achievable cumulative rewards for a current state $s_t$ with action $a_t$.} %, \nj{which estimates the maximum achievable cumulative rewards for a current state $s_t$ with action $a_t$,}  . %Q-function is a function that expects how much cumulative rewards are obtained from this state when a specific action is taken. 
In DQN, %the computational complexity increases exponentially because 
the output of the Q-function is an one-hot vector form of the discrete action space, \nj{which prohibits DQN from being} applied to continuous action space environments.

% DQN에서는 Atari2600 task 학습 환경에서 action-value function($q(s, a)$; Q-function)을 Deep Neural Network로 approximation하여 좋은 성능을 보여주었다.\cite{mnih2013playing} Q-function은 주어진 상태에서 특정한 action을 취했을 때 얼마의 기대 보상값을 얻을지를 표현하는 함수이다. DQN에서 Q-function의 output은 discrete action space의 one-hot vector 형태로 출력되어 가능한 action의 수가 많은 continuous action space에서는 계산량이 기하급수적으로 증가해 적용하기 어렵다는 단점이 있다.
Deep deterministic policy gradient (DDPG) \cite{lillicrap2015continuous} used the Actor-Critic method \cite{peters2008natural} to overcome the disadvantages of DQN in the continuous action space. Trust region policy optimization (TRPO) \cite{schulman2015trust} then used the Kullback-Leibler divergence \cite{kullback1951information} as a constraint to resolve the unstable result of policy update \nj{due from} the \nj{fixed} step size in DDPG. In addition, proximal policy optimization algorithm (PPO) \cite{schulman2017proximal} proposes a clipped surrogate objective function to alleviate the complex \nj{computational requirement} in TRPO.
% Deep Deterministic Policy Gradient(DDPG)\cite{lillicrap2015continuous}에서는 Actor-Critic 방법론\cite{peters2008natural}을 차용해 DQN의 단점을 보완하고 continuous action space에서도 강화학습을 가능하게 하였다. 이어서 Trust Region Policy Optimization(TRPO)\cite{schulman2015trust}에서는 Kullback–Leibler divergence를 constraint로 사용하여 DDPG에서 step size에 따라 Policy Update가 불안정한 부분을 해소하였다. 여기에서 한 발 더 나아가 Proximal Policy Optimization Algorithm(PPO)\cite{schulman2017proximal}에서는 clipped surrogate objective function을 제안함으로써 TRPO에서 요구되는 복잡한 계산 과정을 해소했다.

\section{Environment}
\subsection{Task Specification}
\begin{figure}
	\centering
	\includegraphics[width = 1\linewidth]{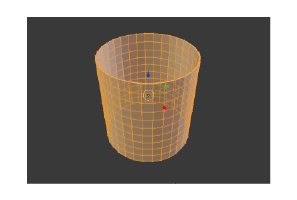}\\
	\caption{Initial state of pot design \nj{drawn in the \textsc{Blender} environment.}}
	\label{pot_design}
\end{figure}

In this study, \nj{to verify the effectiveness of RL in the design process,} we selected the \nj{\textit{cylindrical pot} shown in Fig. \ref{pot_design} } as \nj{a} basic design \nj{for its simple but flexible form.}
%to select the tasks for the most basic form and to verify the design. 
\nj{Starting from a cylinder-shaped pot, the agent tries to maximize the cumulative reward for a specific task by taking actions which is defined as increasing or decreasing the diameter of the pot at different heights.}
The first task \nj{we consider is} pouring as much water as possible in the pot into another cup according to \nj{the} primary purpose of the pot. In addition, we assumed a shaking situation as a second task. Here, the agent tries to keep water from the shaking pot. We named each environment as \nj{`pouring environment'} and `shaking environment', \nj{respectively}. After training each task successfully, we \nj{also show that simultaneously} training both tasks is possible despite these two tasks have conflicting features.
% 본 연구에서, 가장 기본적인 제품에 대해 task들을 선택하고 design 을 검증해보기 위해, pot을 기본 디자인으로 선택하였다. 이 pot의 경우, 다른 물컵으로 물을 따르는 데에 기본 목적이 있기 때문에, 첫번째 task의 경우 다른 물컵으로 물을 따르는 task 를 선정했다. 또한, 두 번째 task로는 흔들리는 상황에서 물이 밖으로 많이 흐르지 않는 task를 선정하였다. 이 두 task는 기본적으로 특징면에서 상충되는 면이 있기 때문에, 본 실험에서는 각각의 task를 기반으로 학습을 시킨 후, 이 두 task를 모두 만족시킬 수 있도록 학습이 가능하다는 것을 보일 예정이다. 각각의 environment를 tilting environment와 shaking environment 로 정의한다.

\subsection{3D Simulation Environment}
The existence of simulation environment \nj{has been} one of the main reasons \nj{for the success of} deep reinforcement learning. Mnih et al. \cite{mnih2013playing} was able to find the optimal policy by training in the \textsc{Atari2600} game environment through Arcade Learning Environment \cite{bellemare2013arcade}. The role of the simulation is also important in \nj{our} task. % we are going to try out in this paper. 
Without simulation, we have to repeat the inefficient process of designing the pot each time, outputting the product, and experimenting.
% Deep Reinforcement Learning이 성공할 수 있었던 이유 중 하나는 시뮬레이션을 할 수 있는 환경이 있었기 때문이다. \cite{mnih2013playing}에서는 Arcade Learning Environment(ALE)\cite{bellemare2013arcade}를 통해 Atari2600 게임 환경에서 traing을 반복하여 최적 정책을 찾아낼 수 있었다. 우리가 이 논문에서 시도하려고 하는 과제에서도 마찬가지로 시뮬레이션의 역할이 중요하다. 만약 시뮬레이션을 할 수 없다면 매번 pot을 디자인하고 제품을 출력하여 실험을 하는 비효율적인 과정을 반복해야만 한다. 

We used an \nj{open-source} 3D modeling tool, \nj{\textsc{Blender}}, to construct a reinforcement learning environment. The reason for using the \nj{\textsc{Blender}} is that it allows fluid simulation through embedded \textit{particle systems} and can control and output all \nj{the available} information \nj{on} the environment via python scripts. The initial model of the pot implemented with the \textsc{Blender} is \nj{shown in} Figure \ref{pot_design}.
% 우리는 강화학습 환경을 구성하기 위해 open source 3D modeling tool인 blender를 차용했다. blender를 사용한 이유는 내장된 particle systems를 통해 유체 시뮬레이션이 가능하고 python script를 통해 환경 안의 모든 정보를 control하고 출력할 수 있기 때문이다. blender로 구현한 pot의 초기 모형은 Figure \ref{pot_design}와 같다.

\subsection{Reinforcement Learning Environment} \label{ssec:rlenv}
As described in the above section, to apply each modeling design to reinforcement learning, we need to define the state and action space. To define the state and action, we assigned 11 \nj{control points along the $z$-axis of the pot by dividing $z$-axis into 10 regions.}
%circle groups by dividing the z axis of the pot by 10. 
And each cross sectional circle \nj{corresponding to each control point} consists of 32 points with equal distances. Now the pot consists of $352 \ (=32 \times 11)$ points as we can see in Figure \ref{pot_design}. The action space is defined as \nj{an} 11-dimensional vector that controls \nj{the radii} of 11 circle groups and the state is a vector of $1,056 \ (= 32\times 11 \times 3)$ dimension which \nj{corresponds to} $x,y,z$ coordinates of all points in the pot design.
% 위 section에서 설명했듯, Reinforcement learning 에 각각의 모델링 디자인을 적용시키려면, state 와 action space 에 대한 정의가 필요하다. state 와 action을 정의하기 위해 위의 pot design 에서 z축을 10등분하여 11개의 point group을 만들었다. 이를 통해 이들의 반지름(r) 값을 변경하는 것으로 action 을 정의하였다. state의 경우, pot을 이루고 있는 점들의 좌표로 정의되며, 각각의 group이 포함하는 32개의 점 * 11개의 group * x,y,z 좌표 3 개로 1056개의 state로 정의된다. 

\begin{figure}[t]
	\centering
	\includegraphics[width = 1\linewidth]{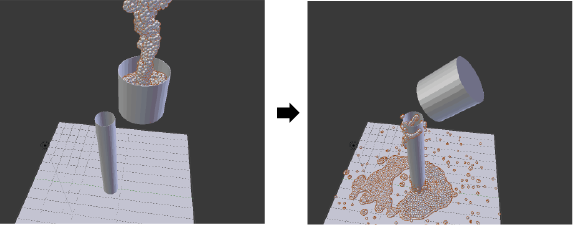}\\
	\caption{Initial step in \nj{the pouring} environment \nj{(left). We simulate the pouring operation by tilting the pot and filling the cup with water (right). The amount of water in the cup is rewarded to the agent which further tries a designing action of increasing or decreasing each diameter of the 11 control points. Note that since the pot and the cup are located with some distance, most of the water spill during pouring operation. }}
	\label{pouring_env}
\end{figure}
As we see in Figure \ref{pouring_env}, in \nj{the pouring} environment, the environment consists of a pot containing water and a cup to receive water. At first, \nj{the agent takes} an action to change the design of the pot from \nj{the} initial state. The environment simulates a step process and then measures a reward. The step proceeds as follows. With a certain amount of water in the pot, tilt the pot from zero to 130 degrees in the \nj{direction of $x$-axis} for 2 seconds. At this time, the amount of water in the cup is measured as a reward.
% Tilting environment의 경우 환경은 물이 담겨있는 pot, 물을 받는 cup으로 구성되며, 유체 시뮬레이션을 통해 해당 단계(step)에서의 행동에 대한 reward를 측정한다. reward는 유체 시뮬레이션을 통해 컵에 담겨지는 물의 양으로 정의하였다. 한 번의 step은 다음과 같이 진행된다. pot에 일정량의 물이 담겨있는 상태에서 pot을 2초 동안 y축 방향으로 130도 만큼 기울인다. 이 때 cup에 담긴 물의 양을 reward로 측정한다. 이 환경에서의 action은 pot의 각각의 point group의 radius를 바꾸는 것으로 정의된다.(앞에서 action 서술했으므로 생략)

\begin{figure}[t]
	\centering
	\includegraphics[width = 1\linewidth]{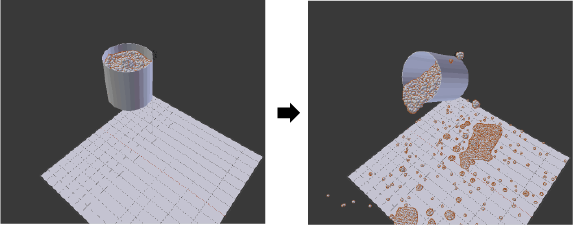}\\
	\caption{Initial step in \nj{the} shaking environment \nj{(left). We simulate the shaking operation by abruptly tilting the pot from 70$^\circ$ to -70$^\circ$. The remaining water in the pot acts as a reward in this environment.} }
	\label{shaking_env}
\end{figure}
In \nj{the} shaking environment \nj{shown in} Figure \ref{shaking_env}, the environment only consists of a pot containing water unlike \nj{the} pouring environment. Most of the \ms{step} processes are similar to the pouring environment except for the simulation step process. With a certain amount of water in the pot, the environment shakes the pot for \ms{thirteen} %two
seconds from $-70$ to 70 degrees in the \nj{direction of $x$-axis}. At this time, the amount of water remaining in the pot is measured as \nj{a} reward.
% Shaking environment의 경우, 환경은 물이 담겨있는 pot으로 구성되며, 유체 시뮬레이션을 통해 해당 step에서의 행동에 대한 reward를 측정하였다. reward는 유체 시뮬레이션 이후에 pot에 남아있는 물의 양으로 정의하였다. 한번의 step은 다음과 같이 진행된다. pot에 일정량의 물이 담겨있는 상태에서 pot을 2초동안 양쪽으로 70도씩 기울며 흔든다. 이 때 pot에 남은 물의 양을 reward로 측정한다. 이 환경에서의 action은 위와 같이 pot의 각각의 point group의 radius를 바꾸는 것으로 정의된다.

\begin{figure*}[h]
	\centering
	\includegraphics[width = 1\linewidth]{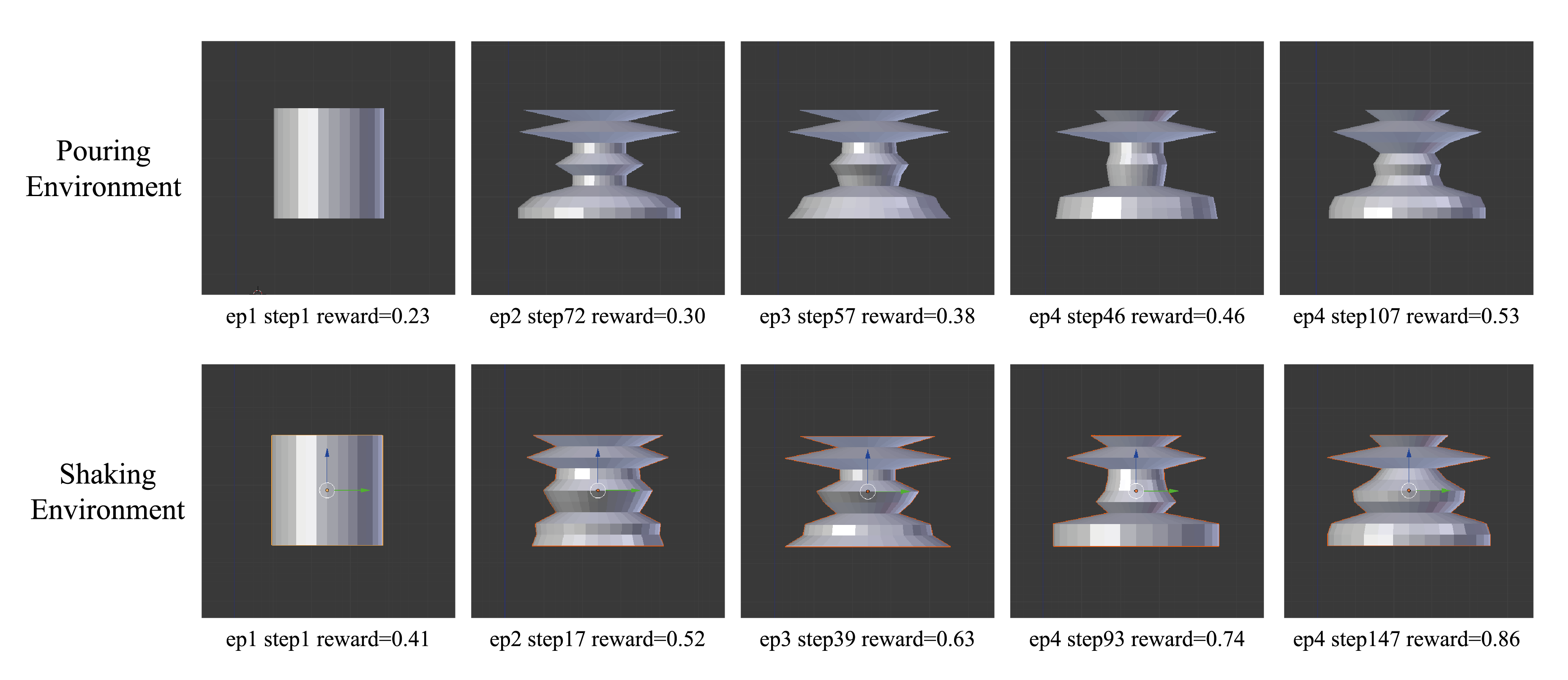}\\
	\caption{Changes in the model during learning. The first row shows the learning process in \nj{the pouring} environment, and the second row shows the learning process in shaking environment. The final image of each raw is when the learning has reached its best performance.}
	\label{figure_change}
\end{figure*}
\section{Experiments}
In this section, we will show that task-oriented design is possible through experiments not only in the tasks mentioned in the previous section, but also in a multi-task environment \nj{where the goal is a hybrid of the two tasks}. And then, we will analyze how the computer understands \nj{a} task and designs \nj{a} model by examining i) \nj{the rewards during learning quantitatively} and ii) \ms{the result of the} %\nj{resulting} 
pot design \nj{qualitatively}.
% In this section, 우리는 먼저 각각의 environment에 대한 실험 결과와(Section 1, 2) 각각을 cross-learning(?) 하여 학습한 결과를 통해 task oriented design이 가능함을 보이고, 이를 해석할 것이다. 이를 각각 i)정량적인 reward 학습의 진행과 ii)정성적인 pot design figure를 분석함으로써, 컴퓨터가 어떤 방식으로 task 를 이해하고 디자인하는지를 분석할 것이다. 

\subsection{Training Model Details}
We used the actor-critic based PPO algorithm for learning experiments. It is composed of actor network and critic network, and is a 3-layer fully-connected network consisting of (256, 128, 64) units. The activation function is tanh (tangent hyperbolic) function. Each network receives 352 points as input that make up the pot. The actor network outputs values corresponding to 11 action spaces defined by \ref{ssec:rlenv} from the gaussian distribution, and the critic network computes the expected cumulative value from current state. For the stability of the learning, the scale of the action is limited to a value between 0.5 and 1.5 times the initial value. We used Adam optimizer as an optimizer and learning rate is 0.0007.
% 실험의 학습에는 actor-critic 기반의 PPO 알고리즘을 사용하였다. 학습 네트워크 actor network와 critic network로 각각 구성되어 있으며 (256, 128, 64) unit으로 이루어진 3-layer fully-connected network이다. Activation function은 tanh(tangent hyperbolic) function을 사용하였다. 각 네트워크는 pot을 구성하는 352개 point를 input으로 받으며 actor network는 \ref{ssec:rlenv}에서 정의한 11개의 action space에 해당하는 값을 gaussian distribution으로부터 뽑아내고, critic network는 현재 state에서의 value function 값을 출력한다. 이때 학습의 안정성을 위해 action인 반지름의 스케일을 초기값 기준 0.5배에서 1.5배 사이의 값으로 제한했다.  Optimizer는 Adam optimizer, learning rate은 0.0007로 사용하였다. 

\subsection{\nj{Pouring} Environment}
In the pouring environment, the pot performs a task of pouring water into a narrow cup \nj{located} a certain distance away \nj{as shown in Figure \ref{pouring_env}}. If the amount of water in the cup is $n_{cup}$, and the total amount of water is $n_p$, the reward is defined as %Eq \ref{reward_pour}.
% Tilting environment에서 pot은 일정 거리에 떨어진 좁은 cup에 물을 따르는 task를 수행하게 된다. cup에 들어간 물의 양을 $n_{cup}$,  물의 총량을 $n_p$라고 할 때 reward 는 Eq.\ref{reward_tilt} 같이 정의된다.
\begin{gather}
reward_{pour} = n_{cup}/n_p.
\label{reward_pour}
\end{gather}
Through this, we designed \nj{an} experiment to get the maximum amount of water to the cup when tilting the pot. We used PPO as the \nj{reinforcement} learning algorithm. In general, \nj{millions to billions} of steps are needed \nj{for a} reinforcement learning model \nj{to converge}. In our experiment, however, we only used 1,000 steps for \nj{training} due to the computational bottleneck of the \textsc{blender} simulation. We separated the 1,000 steps into 5 episodes and initialized the pot design every 200 steps to prevent sub-optimality that might occur in reinforcement learning. Through this, we encouraged the agent to make optimal modeling.
% 이를 통해 pot을 tilt할 때 cup에 최대한의 물을 받도록 학습을 진행하였다. 학습 알고리즘의 경우 PPO를 이용하였다. 일반적으로 강화학습을 진행할 때 모형이 수렴에 이르기까지 적게는 100만 번에서 많게는 10억번의 step을 학습하는데, 우리 실험에서는 blender simulation의 계산 bottleneck이 있어 한 번의 학습 동안 1000번의 timestep을 데이터로 활용하였다. 그리고 1000번의 timestep을 5번의 episode로 분리해 매 200 step마다 환경을 초기화시켜 reinforcement learning에서 일어날 수 있는 sub-optimality를 방지하도록 하였다. 이를 통해, agent가 optimal한 modeling을 하도록 encourage하였다.

\subsubsection{Quantitative Analysis}
\begin{figure}
	\centering
	\includegraphics[width = 1\linewidth]{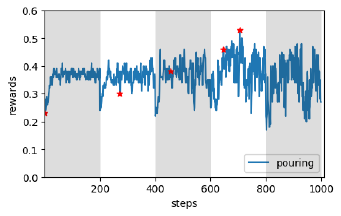}\\
	\caption{Reward Graph in Pouring Environment. \jy{Each red points on the graph represent steps that shown in the first row of Figure \ref{figure_change}.}}
	\label{chart_pour}
\end{figure}

Figure \ref{chart_pour} shows the overall reward rise during the training. 
%\nj{At the top of Figure \ref{figure_change}, we show the five designs obtained during training.} 
\nj{At the} initial state, the reward \nj{is quite low} %loss 
due to the \nj{distance gap in the $x$-direction between the tip of the pot and the center of} the cup. \nj{As the training proceeds, the reward increases}. The valley for each episode \nj{in} Figure \ref{chart_pour} \nj{(at 1, 201, 401, 601, and 801 steps)} shows that it starts again from the initial state so that it deviates from the sub-optimality and shows a slight improvement in reward \nj{as the learning progresses.} %based on the learned network at the beginning of this episode can see. 
Compared to 23\% of water in \nj{the} cup in the initial state, we can see the improvement in performance by containing 53\% of water at the end of the training. The final image in the pouring environment of Figure \ref{figure_change} is what the pot design would look like when it got a reward 0.53 at episode 4 step 107.
% Table.\ref{table_tilting} 과 Figure.\ref{chart_pour}는 학습 도중의 전반적인 reward 상승을 보여준다. initial state에서 pot과 cup간의 거리때문에 발생하는 reward loss가 학습이 진행됨으로써 점점 감소하는 것을 볼 수 있다. Figure.\ref{chart_pour}에서 보이는 episode마다의 valley는 sub-optimality에서 벗어나도록 initial state로부터 다시 시작하는 모습을 보여주며, 이 episode의 시작 부분마다 학습된 network를 바탕으로 조금씩 reward의 향상을 보여주는것을 볼 수 있다. initial state에서 cup에 23%의 물이 담겼던 것에 비해 학습이 끝났을 때 53%의 물이 담긴 것을 통해 성능 향상을 확인할 수 있다. 이를 통해, initial state에서 23\% 의 물밖에 cup에 담기지 않았다면, 학습이 끝났을 때는 53\%의 물을 담아낼 수 있을 정도의 성능 향상을 보여주고 있다. Figure.\ref{figure_change}의 Tilting environment의 마지막 이미지는, algorithm이 0.53 reward를 얻었을 때의 모습이며, episode4 step 107에 기록하였다.

%\begin{table}[!ht]
%  \caption{Rewards Table in Tilting Environment}
%  \label{table_tilting}
%  \begin{tabular}{ccc}
%    \toprule
%    Episode&Steps&Rewards\\
%    \midrule
%    1 & 1 & 0.23\\
%    2 & 72 & 0.30\\
%    3 & 57 & 0.38\\
%    4 & 46 & 0.46\\
%    4 & 107 & 0.53\\
%  \bottomrule
%\end{tabular}
%\end{table}

\subsubsection{Qualitative Analysis}
If you look at the models created by \ms{the deep reinforcement learning} algorithm, you can see which tasks \ms{the agent} %you
 want to perform in each step. In Figure \ref{figure_change}, \nj{the} first row shows how the model trained from the \nj{pouring} environment changes. In the initial state, \nj{it inevitably fails to aim correctly since the distance from the pot to the cup is far as can be seen in Figure \ref{pouring_env}.} Our agent solve this problem by shaping pot design \nj{such that} the center is narrow and the head and \nj{the} bottom are wide, which controls the acceleration of the fluid. \ms{After this,} % When this initial aiming problem is resolved,
 the algorithm passes through \nj{exploration} steps to maximize \nj{the reward (\ref{reward_pour})}. Also, we can see that the head area is resized to maximize the reward.
% Algorithm 에 의해 생성된 model들을 살펴보면, 단계별로 어떠한 task 를 수행하려 하는지 알 수 있다. Figure\ref{figure_change}에서 first row는 tilting environment에서 학습된 model이 어떻게 변하는지를 보여주고 있다. initial state에서 pot부터 cup까지의 거리가 멀기 때문에 필연적으로 발생하는 조준 실패는 training 초반 호리병과 같은 구조를 만들어 유체의 가속을 조절하면서 문제를 해결한다. 이를 통해 가운데가 잘록하고 머리와 바닥부분이 뚱뚱한 구조를 만들어나간다. 이러한 초반 aiming 문제가 해결이 되면, 알고리즘은 Eq.\ref{reward_tilt}를 최대화 시키기 위해 structure를 smoothing하는 과정을 거친다. 또한 head 영역의 크기를 조절하여, reward를 극대화 시키려는 모습을 볼 수 있다.  

\subsection{Shaking Environment}
In the shaking environment, the pot \nj{is designed to shed as little water as possible} in the \nj{environment of shaking the pot}. %cup.
If the \nj{initial amount of water in the pot is $n_p$, and the final} amount of water \nj{after shaking} is \ms{$n_{pot}$}, %$n_{cup}$
\nj{the} reward is defined as
% Shaking environment에서 pot은 흔들리는 컵에서 물을 최대한 흘리지 않는 task를 수행하게 된다. pot에 남아있는 물의 양을 $n_p$, 물의 총량을 $n_{cup}$ 이라고 할 때 reward는 Eq.\ref{reward_shake}와 같이 정의된다.
\begin{gather}
reward_{shake} = \ms{n_{pot}}/n_p.
% reward_{shake} = n_{cup}/n_p
\label{reward_shake}
\end{gather}
Through this, we trained \ms{the pot to} %that when the pot is shaken, it can
\nj{conserve} the maximum amount of water in the \nj{pot}. %In the learning environment, 
We used the PPO algorithm like the \jy{pouring} environment and \ms{trained the model for 5 episodes of 200 steps each.} %5 episodes was learned by 200 steps each.
% 이를 통해 pot이 shake될 때, cup에 최대한의 물을 남길 수 있도록 학습하였다. 학습 환경의 경우 tilting environment와 같이 PPO algorithm을 이용하여, 5번의 episode로 각각 200step씩 학습하였다. 

%\begin{table}[!ht]
%  \caption{Rewards Table in Shaking Environment}
%  \label{table_shaking}
%  \begin{tabular}{ccc}
%    \toprule
%    Episode&Steps&Rewards\\
%    \midrule
%    1 & 1 & 0.41\\
%    2 & 17 & 0.52\\
%    3 & 39 & 0.63\\
%    4 & 93 & 0.74\\
%    4 & 147 & 0.86\\
%  \bottomrule
%\end{tabular}
%\end{table}

\begin{figure}[t]
	\centering
	\includegraphics[width = 1\linewidth]{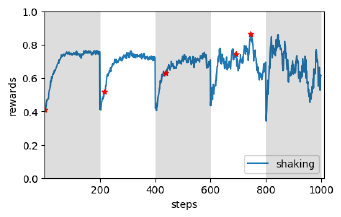}\\
	\caption{Reward Graph in Shaking Environment. \jy{Each red points on the graph represent steps that shown in the second row of Figure \ref{figure_change}.}}
	\label{chart_shake}
\end{figure}
\begin{figure*}[h]
	\centering
	\includegraphics[width = 1\linewidth]{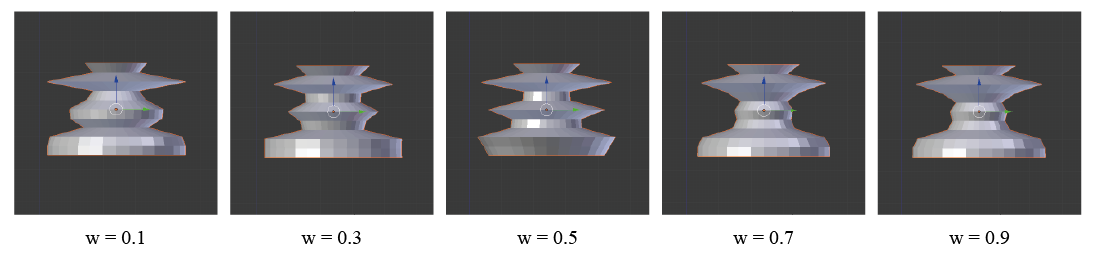}\\
	\caption{Final models in cross-learning environment. Each images represent the best performance in five weight parameter with 0.2 difference from 0.1 to 0.9. \jy{When the weight parameter is close to 0, the learning is focused on solving shaking environment. On the contrary, the higher weight parameter means that the result of learning will be similar to the solution of the pouring environment.}}
	\label{figure_crosslearning}
\end{figure*}
\subsubsection{Quantitative Analysis}
Figure \ref{chart_shake} shows the \nj{trend of reward as the learning progresses} in the shaking environment. The \nj{reward tends} %maximum reward of the episode tends
to decrease slightly after \ms{reaching} \nj{a saturation level} %a certain level 
in the first three episodes, which indicates that the \ms{agent} %network 
\nj{has stuck} in the local minimum during training. However, by solving the local minimum problem after episode 4 \nj{through exploration}, the maximal reward \ms{increases.} %shows an increase.
As a result, \nj{compared to the initial state which saves only 41\% of the water}, at the end of the training, the \nj{resultant pot is able to keep}  86\% of the water. The \nj{bottom right} image of the shaking environment in Figure \ref{figure_change} is when the algorithm gets a reward of 0.86, as recorded in episode 4 step 147.

\subsubsection{Qualitative Analysis}
Looking at the changes in the models generated by \ms{the} reinforcement learning, you can see how the network is trained to protect water. As you can see from \nj{the bottom row of} Figure \ref{figure_change}, the bottom part of the pot becomes larger and larger to keep as much water as possible, and the structure is good for storing water. As training proceeds, it was difficult to store \nj{water} in the lower part, and the training progressed with a double tube structure. Commonly, there is a barrier structure in the upper part to prevent \nj{the pot from} splashing by water shaking. Consequentially, we were able to confirm that when the simulation was carried out, it was trained to keep the bouncing water as much as possible in the pot from the large swing of \nj{$\pm$70} degrees.

\subsection{\nj{Hybrid}-Learning}
In \nj{hybrid}-learning, we examined the possibility of pot design that can perform both contradictory tasks. To do this, we defined a new reward which is \nj{a} weighted sum of the reward in the pouring environment and the reward in the shaking environment as follows:

\begin{gather}
reward_{hybrid} = w \cdot reward_{pour} + (1 \textendash w) \cdot reward_{shake}
\label{reward_crosslearning}
\end{gather}
In this equation, $w$ is a \nj{weight parameter} between 0 and 1. We experimented how the algorithm \nj{interprets} each task according to five $w$ values \nj{in $\{0.1, 0.3, 0.5, 0.7, 0.9 \}$.} % with 0.2 difference from 0.1 to 0.9.

\subsubsection{Quantitative Analysis}
\begin{table}[!h]
  \caption{\nj{Maximum Rewards} in Hybrid-Learning}
  \label{table_cross}
  \begin{tabular}{c|cc|ccc}
    \toprule
    Weight&Episode&Step&Pour&Shake&\nj{Hybrid}\\
    \hline
    \midrule
    0.1 & 4 & 146 & 0.32 & 0.87 & 0.82\\
    0.3 & 5 & 69 & 0.43 & 0.83 & 0.71\\
    0.5 & 5 & 51 & 0.48 & 0.80 & 0.64\\
    0.7 & 4 & 107 & 0.55 & 0.71 & 0.60\\
    0.9 & 4 & 107 & 0.53 & 0.71 & 0.55\\
  \bottomrule
\end{tabular}
\end{table}

Table \ref{table_cross} indicates how much reward is obtained for \nj{different weight parameters}. Each episode and step \nj{indicates} the time when the maximum \nj{hybrid} reward was achieved \nj{for the corresponding weight}, and the \nj{three rewards (pouring, shaking, and hybrid) are the corresponding rewards at the time.}
%the maximum reward obtained from each training. 
When $w$ is 0.1, according to \nj{(\ref{reward_crosslearning})}, we can see that the shaking environment has more weight on training. As \nj{$w$} increases, training is more focused on the pouring environment. As can be seen in Table \ref{table_cross}, in the pouring environment, a pouring reward of 0.32 was achieved at the point where the \nj{hybrid} reward was largest when $w$ was 0.1. \ms{As \nj{$w$} increased to 0.9, } %And 
the \nj{pouring} reward increased to 0.53, which is the best score of the single pouring environment, %when w was 0.9 
because the algorithm gave more weight to the pouring environment.
Conversely, in the shaking environment, the shaking reward was 0.87 when $w$ was 0.1 and the reward decreased to 0.71 when \nj{$w$} was 0.9. In this way, we showed the \ms{deep RL algorithm} %network
combining the two opposite tasks can train a model that satisfies both tasks.

\subsubsection{Qualitative Analysis}
Figure \ref{figure_crosslearning} shows how the model appears based on the change in \ms{the} \nj{weight} \ms{parameter} $w$. When the value of $w$ is 0.1, there is a water trap structure at lower position, a narrow entrance, and a barrier structure below the entrance like the model designed in the \nj{pure} shaking environment. This shows that the training is focused on the shaking environment and trained to maximize water \ms{in the pot.} %saved. 
On the other hand, when $w$ is 0.9, we can see that the model has trained to create a smooth line \nj{in the middle} like the model designed in the pouring environment and flows the water as easily as possible. In the case of the third model \nj{of  $w = 0.5$}, which performs the two tasks in the most balanced way, we can see that the model design maintains all of these features. Though there \nj{exists} a storage part in the middle influenced by the shaking environment, it has a tendency to minimize the water remaining in the pot through the narrowing structure from the bottom to the top which resembles the design of the pouring environment.

\subsection{Contribution in Design Process}
\begin{figure}
	\centering
	\includegraphics[width = 1\linewidth]{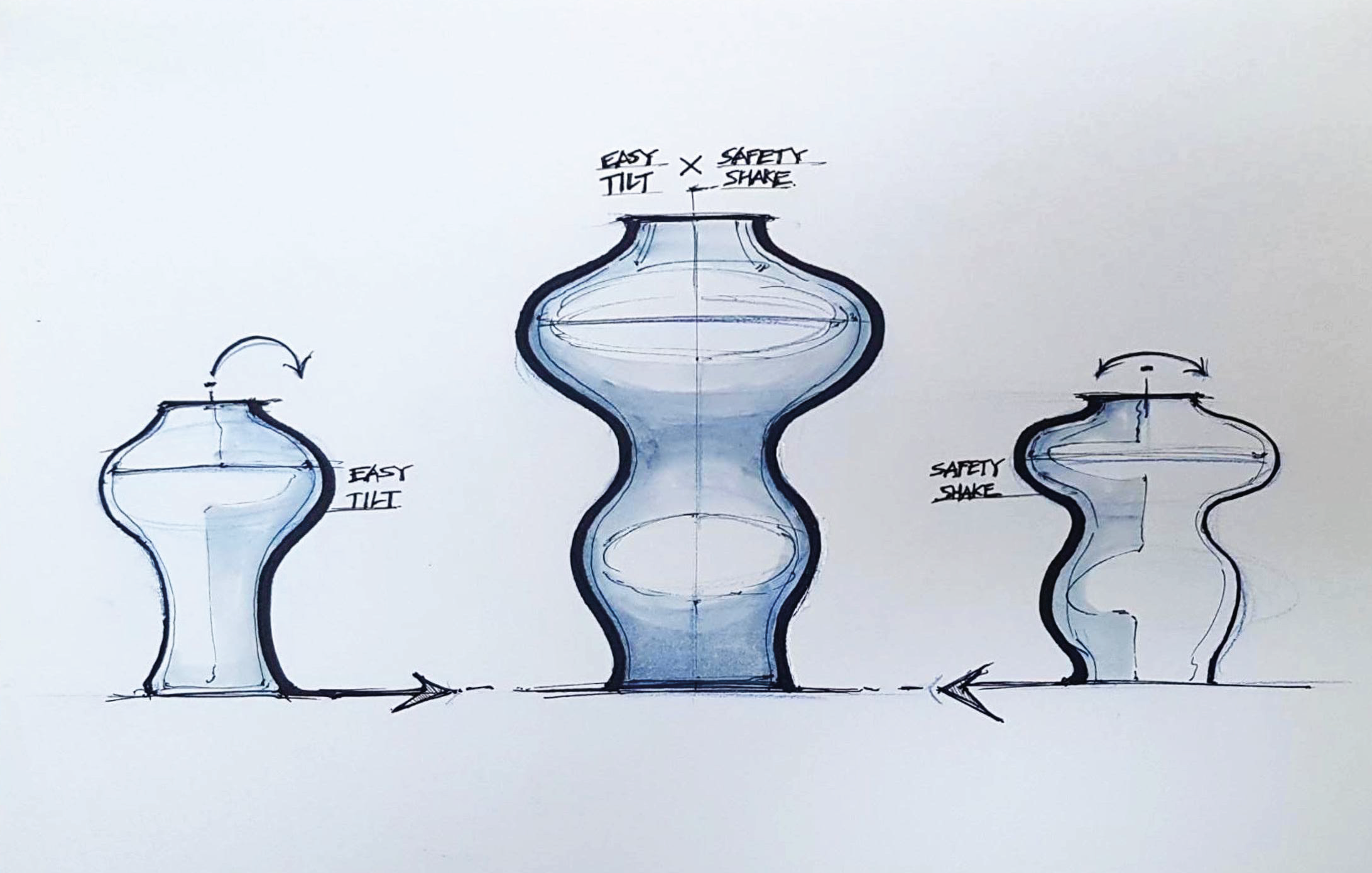}\\
	\caption{The sketch of \nj{a} pot based on \nj{the} computer-generated form. The main concept of design was \nj{Chinese} pot. While designing, \nj{the} product designer get a reference about \nj{the} morphological concept \nj{which is applied to his} actual product design concept.}
	\label{designer_pot}
\end{figure}

Since \nj{the computer-designed products from this experiment does not consider either usability or aesthetics} other than \nj{the given objectives}, it will be necessary to design a creative part based on the generated form. Therefore, in order to verify the validity of the current methodology, we invite\ms{d} a product designer to \nj{sketch designs} through \nj{the} computer-generated design. The tester chose \nj{the} main design concept as a \nj{Chinese} pot. \nj{He} sketched the new \nj{Chinese} pot \nj{using} the \nj{characteristics} of \nj{the} generated form. The design sketch result is shown in Figure \ref{designer_pot}.

After sketching session, we had a short interview with \nj{the} product designer to get pros and cons about this methodological concept. The \nj{designer} commented that \textit{`this methodology is highly useful when designers should make a product from a task-based concept'}. Also, he found out that \textit{the output form is quite similar to the common sense of the pot designed for similar tasks}. As so, \nj{he} believes that the results of this study will be a good reference as a task-based study, which will increase the reliability of the results produced by the designer. However, he has worried that significant features of \nj{the} output can be lost because of the extreme morphological tendency of the computer-generated design.

\section{Conclusions}
Through this study, we \ms{have shown} %show
that task-oriented design using deep reinforcement learning is possible for a \nj{specific task whose objective can be well defined.} 
%obtained through a problem. 
It is shown that the task %success rate 
\nj{objective function} of the modified design through \ms{the deep reinforcement learning} 
is significantly higher than \ms{that of} %in 
the basic form, which \ms{indicates} %shows
that the computer succeeded in designing the task-oriented model. By using this methodology, designers and researchers will be able to apply task-based form research before they move to creative parts of product design process. In addition, the proposed methodology is highly efficient, because it is possible to study morphology within 20 hours at low-cost, \nj{achieving} a high understanding of the task.

However, in the present learning, since an action space is used in which a radius of each point layer is simply changed, there is a limit to an aesthetic or complex design that can be used in real life as an output. Also, since the reward function for the task is simply designed, the %learning 
limit is shown when the performance reaches a certain level. For this reason, we will need \nj{to design a} more delicate action space as well as a reward function in future works.

\bibliographystyle{IEEEtran}
\bibliography{citation.bib}

\end{document}